\documentclass[11pt]{article}

\usepackage[final]{acl}

\usepackage{times}
\usepackage{latexsym}

\usepackage[T1]{fontenc}

\usepackage[utf8]{inputenc}

\usepackage{microtype}

\usepackage{inconsolata}

\usepackage{graphicx}
\usepackage{xspace}
\usepackage[table]{xcolor}
\usepackage{booktabs}
\newcommand{\mname}{{\sc SAGE}\xspace}
\usepackage{algorithm}
\usepackage{algpseudocode}
\usepackage{listings}
\usepackage{changepage}
\usepackage{makecell}
\usepackage{enumitem}
\usepackage{placeins} 
\usepackage{booktabs}
\usepackage{multirow} 
\usepackage{tabularx}

\usepackage{subcaption} 
\usepackage[most]{tcolorbox}
\usepackage{wrapfig}
\usepackage{amssymb}
\usepackage{hyperref}
\usepackage{url}
\usepackage{xcolor}
\usepackage{xspace}

\usepackage{amsmath}
\usepackage{mathtools}
\usepackage{amsthm}
\usepackage{dsfont}

\title{Beyond Static Rules: Automated Discovery of \\Latent Vulnerabilities in Text-to-SQL}

\author{
    Hanqing Wang$^{1*}$, \
    Yongdong Chi$^{1}$\thanks{\ \ Equal Contribution.}, \
    \textbf{Jian Yang}$^{2}$, \\
    \textbf{Lei Yang}$^{3}$, \
    \textbf{Jiehui Zhao}$^{3}$, \
    \textbf{Yun Chen}$^{1}$, \
    \textbf{Guanhua Chen}$^{4}$\thanks{\ \ Corresponding Author.} \\
    $^1$Shanghai University of Finance and Economics, $^2$Beihang University, \\  
    $^3$Deepexi Technology Co. Ltd., 
    $^4$Southern University of Science and Technology \\
}

\begin{document}
\maketitle
\begin{abstract}
While Large Language Models (LLMs) have achieved remarkable success in Text-to-SQL tasks, their deployment in real-world environments is hindered by latent reliability issues. Identifying these latent weaknesses is critical for building trustworthy database interfaces, yet current diagnostic approaches rely heavily on static, expert-defined rules, which lack the capability for systematic and automated exploration. To bridge this gap, we propose SAGE (Systematic Automated Guided Exploration), a novel framework designed to autonomously uncover latent failure patterns in LLM-based Text-to-SQL generation. Specifically, SAGE generates vulnerability hypotheses for given samples and references a continuously evolving Vulnerability Codex to design targeted perturbations, thereby iteratively verifying and documenting potential defects. Extensive experiments on state-of-the-art open-source LLMs demonstrate that SAGE uncovers a substantial number of failure cases, highlighting the significant fragility of current models. Furthermore, our analysis reveals that the Vulnerability Codex exhibits strong cross-model transferability, indicating that the discovered patterns represent generalized structural weaknesses. Finally, we explore SAGE's potential for remediation. Although preliminary, lightweight fine-tuning on the generated samples yields promising improvements, suggesting a scalable pathway for closing the reliability loop in future work. 

\end{abstract}

\section{Introduction}

Text-to-SQL systems~\citep{androutsopoulos1995natural, li2014constructing, li2024can, yu2018spider} serve as a vital bridge between natural language and structured data, democratizing data access by enabling non-technical users to query complex databases intuitively. With the recent integration of Large Language Models (LLMs)~\citep{pourreza2023din,xie2025opensearchsql,li2023resdsql}, this field has witnessed a paradigm shift in semantic understanding and execution accuracy. 

Systematically identifying latent failure modes is the prerequisite for enhancing this reliability. However, current evaluation paradigms predominantly rely on static benchmarks (e.g., Spider~\cite{yu2018spider}, BIRD~\cite{li2023bird}) or manual error analysis~\cite{chang2023drspider,SpiderDK,SpiderSyn, SpiderGen,ADVETA}. 

These approaches are inherently constrained by their reliance on human priors: they primarily verify model robustness against predefined challenges~\cite{zhang2024benchmarking,deng2021structure,AmbiQT}, but cannot provide a systematic exploration of potential or previously unidentified vulnerabilities\footnote{Throughout this paper, ``vulnerability'' refers to robustness and reliability brittleness under semantic-preserving perturbations, rather than database security exploits such as SQL injection.}. As a result, certain failure modes~\cite{saparina2024ambrosia,sahitaj2025utilising,chang2023dr} may remain insufficiently examined and can have a substantial impact on the reliability of Text-to-SQL generation beyond what is captured by fixed heuristics. Consequently, there is a pressing need for an automated framework capable of proactively exploring the model’s failure boundaries.

To bridge this gap, we propose SAGE (\textbf{S}ystematic \textbf{A}utomated \textbf{G}uided \textbf{E}xploration), a framework designed to proactively expose latent vulnerabilities in LLM-based Text-to-SQL generation. Unlike passive verification methods~\cite{chang2023drspider,SpiderDK,SpiderSyn,SpiderGen,ADVETA}, SAGE orchestrates an active discovery process. Specifically, it first formulates specific vulnerability hypotheses for a target sample, and then references a continuously evolving Vulnerability Codex, a dynamic repository of generalized error archetypes, to synthesize targeted perturbed samples. This allows SAGE to systematically verify potential weaknesses and iteratively refine its understanding of the model's failure boundaries, transforming the discovery of vulnerabilities from a static manual task into an autonomous, self-improving cycle.

\vspace{10pt}
Our main contributions are as follows:
\begin{itemize}
    \item We propose \mname, a framework that autonomously evolves a \textit{Vulnerability Codex} to guide exploration. It outperforms static expert-curated baselines, achieving average VER margins of 26.33\% and 18.66\% on BIRD and Spider across different models, respectively. These results indicate the effectiveness of our dynamic, automated strategy compared to fixed manual heuristics.
    
    \item We conduct a systematic vulnerability assessment of LLMs in Text-to-SQL. Our evaluation across diverse models on BIRD and Spider yields average Vulnerability Exposure Rates (VER) of 76.98\% and 58.45\%, respectively. These figures provide empirical evidence of the prevalence of latent vulnerabilities within current Text-to-SQL systems.

    \item Beyond detection, we take an initial step towards establishing a reliability loop. Lightweight fine-tuning on 1.5k samples partially alleviates the model's defects. These findings, though preliminary, shed light on a scalable pathway for model improvement and validate the potential of using \mname for vulnerability remediation.\footnote{Our code is available at \url{https://github.com/sustech-nlp/SAGE}.}
\end{itemize}

\section{Related Work}

\subsection{LLM-based Text-to-SQL Generation}
With the rapid advancement of LLMs, LLM-based approaches to Text-to-SQL have demonstrated significantly enhanced transferability and reasoning capabilities compared to traditional rule-based systems~\cite{mahmud2015rule,lyu2020hybrid}, neural models~\cite{choi2021ryansql,xu2017sqlnet}, and pretrained models~\cite{li2023resdsql,yin2020tabert}, propelling the field into a new stage~\cite{hong2024next, Survey}. Current methodologies primarily follow two paradigms: prompt engineering and fine-tuning. Prompt engineering methods, typically utilizing closed-source LLMs, enhance generation quality through techniques such as task decomposition~\cite{pourreza2023din,xie2025opensearchsql,chi-etal-2025-pi}, chain-of-thought (CoT) reasoning~\cite{xu2024chain,liu2025uncovering}, and multi-agent collaboration~\cite{xia2024r,deng2025reforce}. Conversely, fine-tuning methods leverage open-source LLMs to reduce inference costs while improving privacy, controllability, and stability~\cite{li2025omnisql,qu-etal-2025-share,guo-etal-2025-sqlforge,li2024codes,sheng2025base,cheng2025sqlord}.

Despite their success, the primary objective of these works is to maximize execution accuracy, optimizing the probability of generating correct SQL queries. A specific line of research within this domain—automated self-correction—shares a superficial resemblance to our work by identifying and fixing errors~\cite{shen2025study,gong2025sqlens,askari2025magic}. However, their fundamental goals and mechanisms differ significantly. These methods employ correction solely as an intermediate step to boost performance on specific samples, often addressing errors in an ad-hoc manner without analyzing the underlying causes~\cite{liu2024epi,pourreza2023din}. In contrast, our work shifts the focus from \textit{performance enhancement} to \textit{systematic vulnerability discovery}. We aim not merely to correct individual instances but to abstract high-level failure patterns and diagnose latent defects in the model's reasoning process, providing generalized insights that go beyond instance-level repair.

\begin{figure*}[ht]
    \centering
    \includegraphics[width=\textwidth]{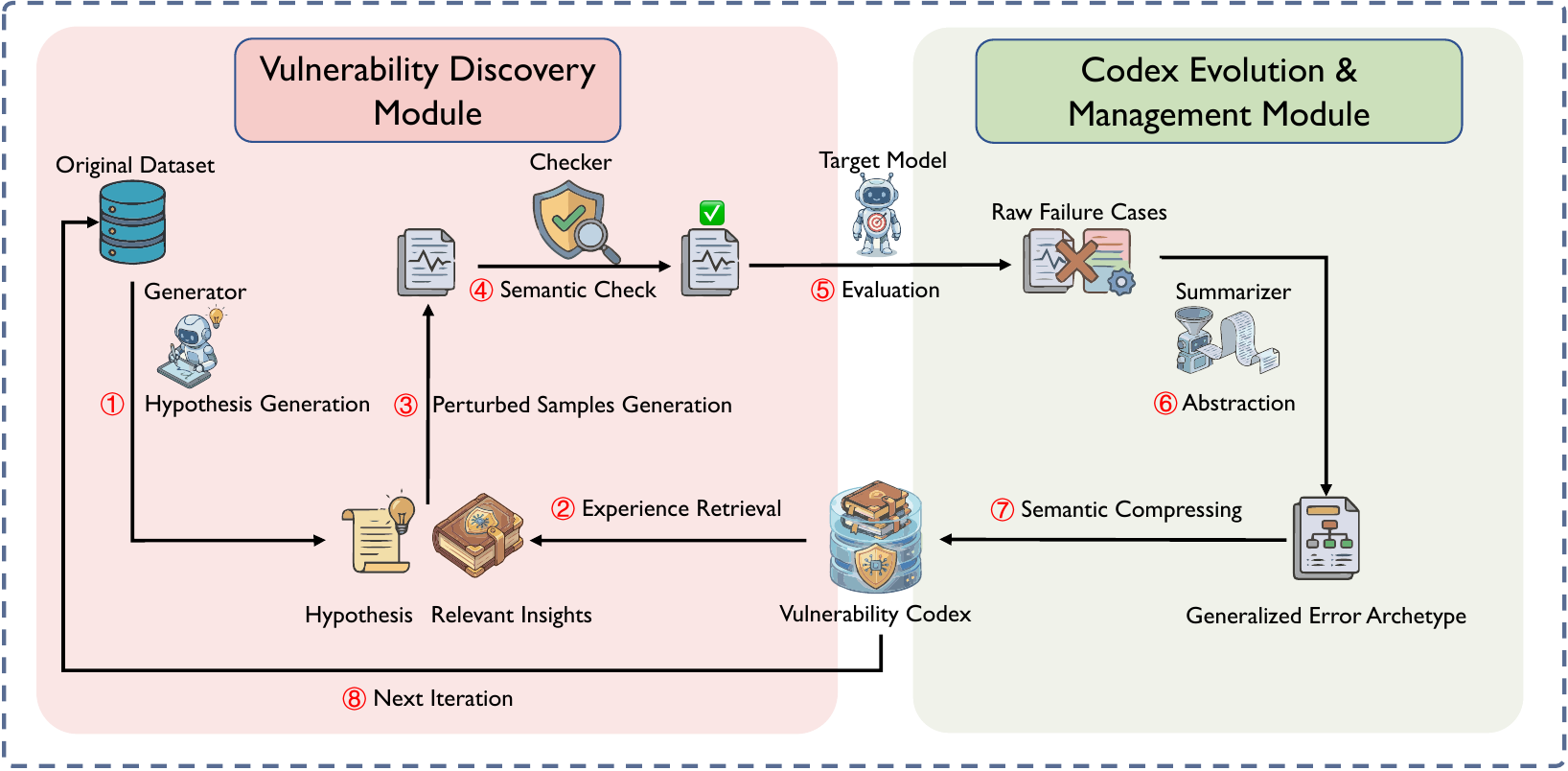}
    \caption{The architecture of the proposed automated vulnerability discovery framework, \mname. The process iterates between two stages: (1) Discovery, where the system utilizes existing insights from the central Vulnerability Codex to generate and verify hard adversarial samples (Steps 1-5); and (2) Evolution, where identified failures are abstracted into generalized archetypes and merged back into the Codex through semantic compression (Steps 6-7). This closed-loop mechanism allows the system to continuously update its strategy in subsequent iterations (Step 8).}
    \label{fig:main}
\end{figure*}

\subsection{Robustness and Brittleness in LLM-based Text-to-SQL Generation}

Prompt-engineering and fine-tuning strategies have continuously advanced Text-to-SQL performance on foundational benchmarks such as Spider~\cite{yu2018spider} and BIRD~\cite{li2023bird}.

However, existing approaches to evaluating model reliability are largely limited to confirming the presence of known failure patterns under specific conditions. Researchers have established robustness benchmarks by applying semantic-preserving perturbations~\cite{deng2021structure, SpiderDK, SpiderSyn, sahitaj2025utilising} or predefined rule-based alterations to schemas and queries~\cite{chang2023dr, SpiderGen}. While valuable, these methods operate within a closed scope defined by human experts—they merely verify whether an LLM succumbs to anticipated error types, such as structural ambiguity~\cite{ding2025ambisql,AmbiQT}, but fail to explore the model's ``blind spots'' outside these heuristics~\cite{saparina2024ambrosia,ADVETA,zhang2024benchmarking}. Although recent efforts pursue automated bias or failure discovery in other LLM settings~\cite{lai2026biasscope}, systematic vulnerability exploration for Text-to-SQL remains largely open.

This highlights the need for an efficient, self-evolving, and automated framework capable of discovering latent failure patterns. 
Unlike previous works, our work aims to systematically expose unknown defect patterns, providing actionable insights for the fundamental robustness of Text-to-SQL systems.

\section{Method}

To systematically uncover latent vulnerabilities in Text-to-SQL systems, we propose \mname, an iterative framework designed to autonomously discover and evolve vulnerability insights. As illustrated in Figure~\ref{fig:main}, \mname operates as a closed-loop system structured around two complementary modules: the \textbf{Vulnerability Discovery Module} (Section~\ref{sec:discovery}), which actively probes the target model through hypothesis-driven perturbations, and the \textbf{Codex Evolution \& Management Module} (Section~\ref{sec:evolution}), which abstracts validated failures into a self-refining \textit{Vulnerability Codex}. The detailed algorithmic procedure is provided in Algorithm~\ref{alg:iterative_discovery_final}.

\subsection{Problem Formulation}
\label{sec:problem_formulation}

We formalize the problem of automatic vulnerability discovery within the Text-to-SQL paradigm. Let $D = \{(q_i, s_i, y^*_i)\}_{i=1}^N$ denote a dataset where $q_i$ is the natural language query, $s_i$ is the database schema, and $y^*_i$ is the ground-truth execution result, on which the target model $M_t$ performs correctly under standard conditions, i.e., $\mathrm{Exec}(M_t(q_i, s_i)) = y^*_i$. This selection ensures that any subsequent failure induced by our framework reflects a genuine blind spot or fragility in the model rather than an inherent lack of reasoning ability.

Our objective is to construct and evolve a \textbf{Vulnerability Codex}, denoted as $\mathcal{V}$, which systematically captures the latent failure modes of $M_t$. We define \mname as an iterative process governed by two mapping functions, $\langle \Phi_{\text{disc}}, \Phi_{\text{evol}} \rangle$, operating over iterations $t=1, \dots, T$.

Let $\mathcal{V}_{t-1}$ denote the Codex state from the previous iteration. At each step $t$:

\paragraph{Discovery ($\Phi_{\text{disc}}$):} The system synthesizes hypotheses and constructs specific perturbed samples to probe $M_t$, yielding a set of raw failure cases $\mathcal{F}_t$ where $M_t$ exposes vulnerability:
\begin{equation}
    \mathcal{F}_t = \Phi_{\text{disc}}(D, \mathcal{V}_{t-1})
\end{equation}

\paragraph{Evolution ($\Phi_{\text{evol}}$):} These raw failures are abstracted, compressed, and integrated into the Codex to produce an updated state $\mathcal{V}_t$:
\begin{equation}
    \mathcal{V}_t = \Phi_{\text{evol}}(\mathcal{V}_{t-1}, \mathcal{F}_t)
\end{equation}

The process continues until the Codex stabilizes or the maximum iteration budget is reached.

\subsection{Vulnerability Discovery Module}
\label{sec:discovery}

This module executes the active probing phase of our framework. It transforms abstract vulnerability hypotheses into concrete, context-aware adversarial samples to expose latent defects. This process follows a coarse-to-fine pipeline: from \textit{Hypotheses} to \textit{Scopes}, and finally to \textit{Sample Generation}.

\subsubsection{Systematic Hypothesis Generation}

Diverging from stochastic perturbation methods, we initiate discovery by synthesizing targeted \textbf{Vulnerability Hypotheses}. A Generator Agent ($M_g$) analyzes each data sample $d_i \in D$ to identify potential vulnerabilities (e.g., ``Ambiguity in column selection'' or ``Misinterpretation of complex nested logic''). This yields a set of candidates $\mathcal{C}_{curr} = \bigcup_{d_i \in D} \{(d_i, h_{i,k})\}_{k=1}^K$, where each $h_{i,k}$ represents a specific conceptual strategy suspected to induce failure (step 1 in Figure~\ref{fig:main}).

To ensure structural grounding, these hypotheses are mapped to specific \textbf{Perturbation Scopes} $\mathbb{S}$ (e.g., logical constraints in the query or schema definitions). In the initial phase ($t=1$), we adopt an efficient exploration strategy by mapping each hypothesis to a single scope; in subsequent iterations, a comprehensive scan is performed by applying hypotheses across all available scopes.

\subsubsection{Experience-Guided Probing \& Verification}

To rigorously verify whether a candidate hypothesis $C_j = (d_i, h_{i,k})$ exposes actual vulnerabilities, \mname references the \textit{Vulnerability Codex} to guide the generation process. We employ an Embedding Model ($M_e$) to retrieve relevant insights $E_j$ from $\mathcal{V}_{t-1}$. This serves as a reference for implementation feasibility rather than content limitation, ensuring that the verification focuses on the hypothesis's validity (step 2 in Figure~\ref{fig:main}).
\begin{equation}E_j = \mathrm{RetrieveTop5}(M_e, h_{i,k}, \mathcal{V}_{t-1})\end{equation}

Guided by the candidate $C_j$, the target scope $s_m \in \mathbb{S}$, and the retrieved experience $E_j$, the Generator $M_g$ constructs a perturbed sample $x_{j,m}$ (step 3 in Figure~\ref{fig:main}):

\begin{equation}
    x_{j,m} = \mathrm{SampleGenerate}(M_g, C_j, s_m, E_j)
\end{equation}

To ensure $x_{j,m}$ preserves the original semantic logic (i.e., the ground truth execution result remains invariant), we introduce a Checker Agent ($M_c$) to validate the qualification of $x_{j,m}$ (step 4 in Figure~\ref{fig:main}). 

\begin{equation}
    \mathrm{IsValid}=\mathrm{ValidCheck}(M_c, C_j, x_{j,m})
\end{equation}

Given the complexity of semantic verification, we validate the Checker Agent's reliability through both human evaluation and cross-model proxy validation (see Appendix~\ref{sec:checker_validation}).

Upon validation, we probe the target model $M_t$ with $x_{j,m}$ to obtain the predicted SQL execution result $y_{j,m}$. A raw vulnerability is identified if the perturbation is valid, but the model execution deviates from the ground truth (step 5 in Figure~\ref{fig:main}):

\begin{equation}
    \mathrm{IsCorrect} = (\mathrm{Exec}(M_t(x_{j,m})) == y^*_i)
\end{equation}

If $\mathrm{IsCorrect}$ is false, the sample reveals a genuine vulnerability, triggering the evolution phase.

\subsection{Codex Evolution \& Management Module}
\label{sec:evolution}

While the Discovery Module exposes individual failures, the Evolution Module functions as the system's cognitive repository, responsible for distilling knowledge. It converts noisy, instance-specific failures into a high-quality, compact \textbf{Vulnerability Codex} ($\mathcal{V}$). This involves two key stages: \textit{Abstraction} and \textit{Refinement}.

\subsubsection{Abstraction via Summarization}

Raw failure cases are often coupled with specific schema entities or query phrasings. To foster generalization, a Summarizer Agent ($M_s$) analyzes the discovered failure tuple $(C_j, x_{j,m}, y_{j,m})$ (step 6 in Figure~\ref{fig:main}). It abstracts the concrete error into a structured \textbf{Generalized Error Archetype} $v_{j}$, which encapsulates a generalized error summary and causal analysis.

\begin{equation}
    v_{j} = \mathrm{VulnerAbstraction}(M_s, C_j, x_{j,m}, y_{j,m})
\end{equation}

These abstracted archetypes are collected into a temporary set $\mathcal{V}_{\text{new}}^{(t)}$:
\begin{equation}
    \mathcal{V}_{\text{new}}^{(t)} = \mathcal{V}_{\text{new}}^{(t)} \cup \{v_j\}
\end{equation}

\subsubsection{Refinement via Semantic Compression}

As the iterative process proceeds, simply accumulating archetypes in $\mathcal{V}_{\text{new}}^{(t)}$ may lead to a redundant repository. To address this, we introduce a \textbf{Semantic Compression} mechanism during the construction of $\mathcal{V}_t$ (Phase 3 in Algorithm~\ref{alg:iterative_discovery_final}).

At the conclusion of each iteration, newly discovered archetypes are merged with the existing Codex (step 7 in Figure~\ref{fig:main}). We perform semantic clustering using $M_e$: if the Euclidean distance between the embeddings of two vulnerability entries falls below a similarity threshold $\tau$, they are identified as redundant and compressed.

Formally, this refinement updates the Codex by:

\begin{equation}
    \mathcal{V}_t = \mathrm{SemanticComp}(M_e, \mathcal{V}_{t-1} \cup \mathcal{V}_{\text{new}}^{(t)}, \tau)
\end{equation}

This refinement step ensures that the Vulnerability Codex $\mathcal{V}$ remains compact and distinct, improving retrieval efficiency and enabling the system to converge towards a representative set of failure modes on Text-to-SQL generation.

\section{Experiment}
\subsection{Setup}
\paragraph{Datasets}

We evaluate \mname using the \textit{Spider}~\cite{yu2018spider} and \textit{BIRD}~\cite{li2023bird} datasets. To rigorously benchmark vulnerability, we define our evaluation set as the specific subset of samples where the target model yields the correct ground-truth execution result, filtering out instances of initial failure. The exact cardinalities of this initially-correct subset $|D|$ for each model/dataset pair are reported in Appendix Table~\ref{tab:reproducibility}.

\paragraph{Models}

For our main experimental evaluations, we select three representative models that span distinct paradigms as our target models: a general-purpose model, Gemma-3~\cite{team2025gemma} (\textit{Gemma-3-12B-it}); a domain-adapted model, OmniSQL~\cite{li2025omnisql} (\textit{OmniSQL-32B}); and a reinforcement-learned model, Inf-rl-qwen~\cite{infly2025infrlqwencoder} (\textit{inf-rl-qwen-coder-32B-2746}). Additionally, we demonstrate the applicability of our approach to powerful proprietary models by evaluating on \textit{GPT-4o}~\cite{hurst2024gpt} (see Appendix~\ref{sec:transfer-sota}). For fine-tuning experiments in Section~\ref{sec:fine-tuning}, we use Qwen2.5-Coder(\textit{Qwen2.5-Coder-7B-Instruct}), considering training cost.

\begin{table*}[ht]
  \centering
  \scriptsize
  \renewcommand{\arraystretch}{1.2}
  \setlength{\tabcolsep}{5pt}
  
  \resizebox{\textwidth}{!}{%
    \begin{tabular}{llcccccccc}
      \toprule
      \multirow{2}{*}{\textbf{Model}} & \multirow{2}{*}{\textbf{Setting}} &
      \multicolumn{4}{c}{\textbf{BIRD}} & \multicolumn{4}{c}{\textbf{Spider}} \\
      \cmidrule(lr){3-6} \cmidrule(lr){7-10}
      & & \textbf{EX} ($\downarrow$) & \textbf{VES} ($\downarrow$) & \textbf{VER} ($\uparrow$) & \textbf{ApD} ($\downarrow$)
        & \textbf{EX} ($\downarrow$) & \textbf{VES} ($\downarrow$) & \textbf{VER} ($\uparrow$) & \textbf{ApD} ($\downarrow$) \\
      \midrule
      
      \multirow{3}{*}{\textbf{Gemma-3}} 
        & Original & 53.65 & 63.75 & --    & --   & 81.53 & 109.35 & --    & --   \\
        & Expert   & 24.64 & 29.38 & 54.07 & 7.73 & 48.36 & 61.83  & 52.16 & 10.95\\
        & \mname  & \textbf{8.34}  & \textbf{9.86}  & \textbf{84.45} & \textbf{5.08} & \textbf{12.38} & \textbf{15.72}  & \textbf{66.32} & \textbf{7.85} \\
      \midrule
      
      \multirow{3}{*}{\textbf{Inf-rl-qwen}}
        & Original & 70.53 & 87.03 & --    & --   & 87.72 & 116.38 & --    & --   \\
        & Expert   & 38.20 & 44.87 & 45.84 & 9.81 & 63.93 & 82.11  & 27.12 & 19.97\\

         & \mname  & \textbf{21.90} & \textbf{26.03} & \textbf{68.95} & \textbf{7.27} & \textbf{56.19} & \textbf{71.05}  & \textbf{48.07} & \textbf{12.93} \\

      \midrule
      
      \multirow{3}{*}{\textbf{OmniSQL}}

        & Original & 63.23 & 74.77 & --    & --   & 81.83 & 110.85 & --    & --   \\
        & Expert   & 30.31 & 36.63 & 52.06 & 7.09 & 48.84 & 63.57  & 40.10 & 10.60\\

        & \mname  & \textbf{14.21} & \textbf{17.06} & \textbf{77.53} & \textbf{5.47} & \textbf{31.82} & \textbf{40.35}  & \textbf{60.97} & \textbf{8.15} \\

      \bottomrule
    \end{tabular}
  }
  \vspace{0.5em}
  \caption{Main results on \textbf{BIRD} and \textbf{Spider} datasets. 
  We report \textbf{EX} (Execution Accuracy), \textbf{VES} (Valid Efficiency Score), \textbf{VER} (Vulnerability Exposure Rate), and \textbf{ApD} (Attempts per Discovery). 
  Arrows indicate whether higher ($\uparrow$) or lower ($\downarrow$) values are preferred. 
  \emph{Original} denotes the standard baseline performance. 
  \emph{Expert} and \emph{\mname} represent vulnerability probing methods via manual rules and our automated framework, respectively. 
  Note that VER and ApD quantify the discovery capability and are not applicable (--) to the static Original baseline.}
  \label{tab:main_results}
\end{table*}

\paragraph{Implementation Details} \label{sec:detector}

We initialize the Vulnerability Codex using a repository of expert-defined vulnerabilities curated from established studies, including Dr.Spider, SpiderSyn, and ADVETA~\cite{chang2023dr, SpiderSyn, ADVETA}. We employ \textbf{Qwen3-32B}~\cite{yang2025qwen3} as the backbone for the Generator, Checker, and Summarizer components, while \textbf{Qwen3-Embedding-4B}~\cite{zhang2025qwen3} serves as the embedding model. Unless otherwise specified, we use $T=3$ iterations, $K=3$ initial hypotheses per sample, Top-5 Codex retrieval, and a semantic compression threshold $\tau=0.1$; the search over a hypothesis stops once a valid failure is exposed or the maximum iteration budget is reached. We follow each model's default chat sampling settings during generation (for Qwen3-32B, $temperature=0.6$, $top_k=20$, and $top_p=0.95$). A consolidated summary of these settings, together with the exact $|D|$ values, is provided in Appendix Table~\ref{tab:reproducibility}. Full details regarding the prompts used for sample perturbation are provided in Appendix~\ref{sec:prompts}. Given the critical role of semantic verification, we provide a dedicated validation of the Checker module in Appendix~\ref{sec:checker_validation}.

For fine-tuning experiments involving \textit{Qwen2.5-Coder-7B-Instruct}, we utilize LoRA for parameter-efficient adaptation. Our training pipeline is built upon the \texttt{llama-factory}\footnote{\url{https://github.com/hiyouga/LlamaFactory}} framework. Comprehensive hyperparameter configurations are listed in Appendix Table~\ref{tab:train_hyper}.

\paragraph{Baselines}
To demonstrate the effectiveness of \mname in discovering latent vulnerabilities on text-to-SQL generation, we introduce such baselines:

\noindent $\bullet$ \textbf{Original} Represents the standard performance of the target model on the unperturbed benchmarks (Spider and BIRD), serving as the reference for initial capability.

\noindent $\bullet$ \textbf{Expert} Due to the absence of direct baselines for automated vulnerability discovery in this specific context, we construct a strong baseline derived from prior work. This method executes vulnerability discovery using a fixed repository of expert-defined patterns (from Dr.Spider, SpiderSyn and ADVETA) without the dynamic update mechanism of our framework. This comparison is designed to highlight the advantage of evolving the Vulnerability Codex versus relying solely on static expert knowledge.

\paragraph{Metrics}
We evaluate model performance using three complementary metrics:

\noindent $\bullet$ \textbf{Execution Accuracy (EX):} EX measures the ratio of correctly predicted SQL programs by comparing their execution results with those of the ground-truth SQL programs on the same database instance.

\noindent $\bullet$ \textbf{Valid Efficiency Score (VES):} 
VES evaluates the efficiency of correctly predicted SQL programs by considering their execution time.
Formally, VES is given by:
$$\mathrm{VES} = \frac{1}{N} \sum^{N}_{n=1}\mathds{1}(n)\cdot R(n)$$,
Where $R(n)$ denotes the ratio between the execution time of the predicted SQL and the gold SQL for the n-th text query. The indicator function $\mathds{1}(n)$ equals 1 if the predicted SQL is correct, and 0 otherwise.

\noindent $\bullet$ \textbf{Vulnerability Exposure Rate (VER):}
This metric quantifies the method's \emph{capability} to uncover latent defects in scenarios where the model typically performs well.
We focus specifically on the set of instances $D$ that the target model solves correctly under standard baseline conditions.
Let $D_{exposed} \subseteq D$ denote the subset of these instances for which our method successfully induces a valid failure.
The VER is defined as:
\[
\mathrm{VER} \;=\; \frac{|D_{exposed}|}{|D|}.
\]
A higher VER indicates a stronger capability of the evaluation framework to penetrate the model's superficial robustness and reveal deep-seated vulnerabilities that remain hidden during standard testing.

\noindent $\bullet$ \textbf{Attempts per Discovery (ApD):}
We introduce this metric to quantify the \emph{efficiency} of capturing valid failure patterns.
Let $m$ denote the total number of interaction attempts, and $n$ denote the number of \emph{distinct, validated discoveries} (deduplicated by root issue).
We define ApD as:
\[
\mathrm{ApD} \;=\; \frac{m}{\max(n,\,1)}.
\]
This ratio reflects the \textbf{discovery efficiency}: a lower ApD indicates a highly efficient process where valid issues are surfaced with fewer interactions. 

\subsection{Main Results}

Table \ref{tab:main_results} presents the comparative evaluation results on the BIRD and Spider benchmarks. We assess model performance under the standard setting (\textit{Original}) against two vulnerability discovery settings: the rule-based \textit{Expert} baseline and our proposed automated framework, \textit{\mname}. The results indicate a substantial performance gap between the standard evaluation and the vulnerability discovery settings across both datasets. On the \textbf{BIRD} dataset, while the evaluated models achieve an average Execution Accuracy (EX) of $62.47\%$ in the \textit{Original} setting, this metric decreases to an average of $14.82\%$ under the \textit{\mname} setting. A similar trend is observed on \textbf{Spider}, where the average EX drops from $83.69\%$ to $33.46\%$ when subjected to our framework. These declines in EX and the corresponding Valid Efficiency Score (VES) suggest that widely used benchmarks may not fully reflect the reliability of models when facing semantically equivalent but adversarial contexts constructed for vulnerability discovery.

Comparing the two discovery methods, \textit{\mname} demonstrates consistent improvements over the \textit{Expert} baseline in terms of both discovery capability and efficiency. Regarding capability, \textit{\mname} exposes a higher proportion of vulnerabilities across both benchmarks. On BIRD, our method achieves an average Vulnerability Exposure Rate (VER) of $76.98\%$, surpassing the \textit{Expert} baseline's average of $50.66\%$. Similarly, on Spider, \textit{\mname} outperforms the baseline with an average VER of $58.45\%$ compared to $39.79\%$. We perform an analysis of discovered vulnerability patterns in Appendix~\ref{sec:error} for a better understanding of the effectiveness of \mname. In terms of efficiency, \textit{\mname} requires fewer interactions to identify valid issues, as evidenced by the Attempts per Discovery (ApD) metric. The average ApD on BIRD decreases from $8.21$ with the \textit{Expert} method to $5.94$ with \textit{\mname}, and on Spider, it decreases from $13.84$ to $9.64$, indicating a more targeted discovery process.

The quantitative advantage of \textit{\mname} can be attributed to its dynamic knowledge refinement mechanism. While the \textit{Expert} approach relies on a fixed set of pre-defined heuristic rules, which limits its discovery scope to anticipated patterns, \textit{\mname} utilizes information from identified vulnerabilities to guide subsequent exploration during the runtime. This feedback loop allows the framework to adaptively target specific weaknesses in the LLMs, resulting in higher exposure rates and improved search efficiency compared to static strategies. This demonstrates the necessity of an evolving, automated framework for comprehensively auditing model robustness.

\begin{table}[t]
  \centering
  \small
  \renewcommand{\arraystretch}{1.15}
  \setlength{\tabcolsep}{4pt} 
  
  \resizebox{\linewidth}{!}{%
    \begin{tabular}{l cc cc}
      \toprule
      \multirow{2}{*}{\textbf{Perturbation Scope}} & \multicolumn{2}{c}{\textbf{VER} ($\uparrow$)} & \multicolumn{2}{c}{\textbf{ApD} ($\downarrow$)} \\
      \cmidrule(lr){2-3} \cmidrule(lr){4-5}
       & \textbf{Score} & $\boldsymbol{\Delta}$ & \textbf{Score} & $\boldsymbol{\Delta}$ \\
      \midrule
      
      Query Only     & 73.75 & -10.70 &  5.54 & +0.46 \\
      Relevant Schema Only       & 51.15 & -33.30 &  8.90 & +3.82 \\
      Irrelevant Schema Only     & 48.12 & -36.33 & 10.84 & +5.76 \\
      
      \midrule
      
      \rowcolor{gray!10} \textbf{\mname{}} & \textbf{84.45} & -- & \textbf{5.08} & -- \\
      
      \bottomrule
    \end{tabular}%
  }
  \vspace{0.2em}
  \caption{
    Ablation study on \textbf{BIRD} with Gemma-3, analyzing the impact of distinct perturbation scopes.
    We report \textbf{VER} (Vulnerability Exposure Rate, higher is better) and \textbf{ApD} (Attempts per Discovery, lower is better).
    \textbf{\mname} targets all scopes simultaneously. 
    $\boldsymbol{\Delta}$ values indicate the performance degradation (lower VER or higher ApD) when restricting the perturbation to a single scope.
  }
  \label{tab:ablation_scopes}
  \vspace{-10pt}
\end{table}

\subsection{Ablation Study: Impact of Perturbation Scopes}

Table \ref{tab:ablation_scopes} investigates the contributions of different input components—Query \& Evidence, Relevant Schema, and Irrelevant Schema—to the vulnerability discovery process. We compare the full \textit{\mname} framework against restricted baselines where perturbations are confined to a single scope.

\textbf{Effectiveness of Individual Scopes.}
The results indicate that perturbations applied to any individual scope are effective in revealing model defects.
Confining the exploration to the \textit{Query \& Evidence} alone yields a high Vulnerability Exposure Rate (VER) of $73.75\%$, confirming that semantic variations in the natural language question are a primary source of model confusion.
Similarly, modifying only the \textit{Relevant Schema} or \textit{Irrelevant Schema} also uncovers a significant portion of failures, with VERs of $51.15\%$ and $48.12\%$ respectively.
This demonstrates that vulnerabilities are distributed across all dimensions of the input context, and each component serves as a critical entry point for probing model's potential vulnerability.

\textbf{Superiority of the Comprehensive Scope.}
While single-scope strategies are valid, \textit{\mname} achieves superior performance by covering the comprehensive input space.
First, regarding \textbf{discovery capability}, \textit{\mname} attains the highest VER of $84.45\%$, outperforming the best single-scope variant by over $10$ percentage points. By not limiting the perturbation to a single dimension, our method can identify a wider range of failure patterns that exist across the query and schema components.

Second, regarding \textbf{discovery efficiency}, \textit{\mname} achieves the lowest Attempts per Discovery (ApD) score of $5.08$, indicating a more efficient search process than any restricted setting.
We attribute this efficiency gain to the data-driven nature of our \emph{Knowledge Refinement} mechanism.
Since the full-scope approach exposes a larger volume and variety of failures (higher VER), it provides richer feedback data for the system to learn from. This allows \textit{\mname} to accumulate more diverse failure patterns and guide subsequent exploration more effectively, thereby accelerating the discovery process compared to searching within a constrained scope.

\section{Analysis}

\begin{figure}[t]
  \centering
  \includegraphics[width=0.9\linewidth]{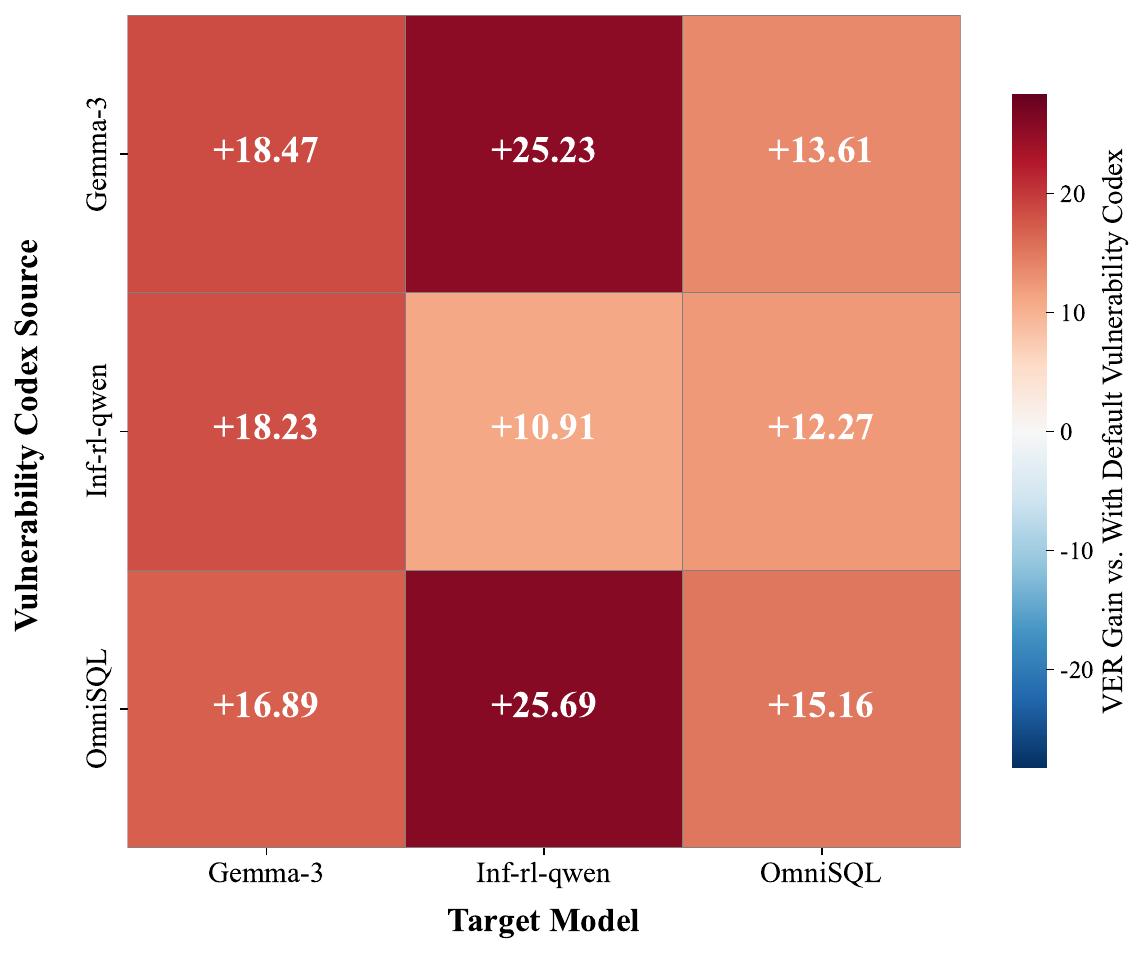}
  \caption{Cross-model strategy transfer performance. The heatmap illustrates the absolute VER gain of Target Models (x-axis) when \mname uses a Vulnerability Codex constructed from different Source Models (y-axis), compared with the \textit{Expert} baseline that uses a static codex. The values are absolute gains over the Expert baseline under the same evaluation framework.}
  \label{fig:cross-model-transfer}
  \vspace{-5mm}
\end{figure}

\subsection{Cross-Model Discovered Vulnerability Generalization Analysis}
\label{sec:transfer_analysis}

Figure~\ref{fig:cross-model-transfer} illustrates the cross-model generalization of \mname by visualizing the VER gains of a Target Model (x-axis) when guided by the vulnerability codex derived from a Source Model (y-axis), measured as absolute gains relative to the \textit{Expert} baseline. Along the main diagonal, where each target model uses its own evolved codex, the self-codex gains range from $+10.91\%$ to $+18.47\%$, showing that dynamically refined strategies consistently outperform manually curated expert rules. We also observe robust cross-model transferability: strategies derived from one model (e.g., OmniSQL) still improve the VER of distinct architectures (e.g., Gemma-3 by $+13.74\%$). This indicates that the discovered vulnerabilities capture structural weaknesses that generalize across models rather than merely reflecting model-specific artifacts.

\subsection{Mitigating Vulnerability with Fine-tuning}
\label{sec:fine-tuning}

The adversarial samples produced by \mname's discovery process constitute a targeted form of data synthesis for LLM training~\cite{zhu-etal-2025-fanno,zhu-etal-2025-tag}. To investigate whether these samples can be transformed into actionable defense signals, we conducted a lightweight mitigation experiment.
Specifically, we selected 1,000 samples from the BIRD training set and employed \mname to discover vulnerabilities on \textit{Gemma-3}, generating 1,580 adversarial samples. These samples—representing high-confidence failure modes or ``hard'' cases—were then used to perform Supervised Fine-Tuning (SFT) on \textit{Qwen2.5-Coder-7B-Instruct}.

As presented in Table~\ref{tab:fine-tuning}, this targeted fine-tuning yields consistent improvements. Under the adversarial evaluation of \mname, the Execution Accuracy (EX) of the model rises from $4.37\%$ to $9.58\%$, representing an absolute gain of $+5.21\%$ and a relative improvement of over \textbf{$119\%$} compared to the base model. On the standard BIRD dev set, EX also improves from $42.63\%$ to $52.87\%$, indicating no observable robustness-general accuracy trade-off in this setting. Simultaneously, Attempts per Discovery (ApD) increases from $3.61$ to $4.59$.

These results are particularly notable given the data efficiency of the approach—using only 1,580 samples to achieve such gains. The concurrent rise in both EX and ApD indicates that the model has not only corrected specific errors but has also become inherently more resilient; it now forces the attacker to search deeper to find a successful perturbation. This suggests that the specific structural weaknesses targeted by our adversarial data have been effectively mitigated, validating the quality of the generated samples.

Crucially, this experiment provides a practical validation of the transferability phenomenon discussed in Section~\ref{sec:transfer_analysis}. By successfully repairing vulnerabilities in \textit{Qwen2.5-Coder} using adversarial patterns discovered in \textit{Gemma-3}, we reconfirm that the flaws exposed by \mname are not model-specific artifacts but generalized linguistic or logic loopholes shared across different LLMs.

Ultimately, this experiment highlights the dual value of \mname. It functions not merely as a diagnostic tool for identifying defects but as a constructive framework capable of ``closing the loop''—enabling researchers to not only pinpoint weaknesses but also leverage them to synthesize robust training data, thereby actively enhancing the reliability of large language models.

\begin{table}[t]
\centering
\small
{
\setlength{\tabcolsep}{4.5pt}
\begin{tabular}{lccc}
\toprule
\multirow{2}{*}{\textbf{Model Setting}} & \multicolumn{2}{c}{\textbf{EX}} & \multirow{2}{*}{\textbf{ApD}} \\
\cmidrule(lr){2-3}
 & \textbf{Adv.} & \textbf{Std.} &  \\
\midrule
Qwen2.5-Coder & 4.37 & 42.63 & 3.61 \\
\addlinespace[0.2em] 
\quad + SFT & \makecell[c]{\textbf{9.58} \\ \small{\textcolor{teal}{(+5.21)}}} & \makecell[c]{\textbf{52.87} \\ \small{\textcolor{teal}{(+10.24)}}} & \makecell[c]{\textbf{4.59} \\ \small{\textcolor{teal}{(+0.98)}}} \\
\bottomrule
\end{tabular}
}

\caption{Mitigation results via adversarial fine-tuning. We fine-tune \textit{Qwen2.5-Coder-7B-Instruct} using 1,580 adversarial samples generated by \mname (targeting \textit{Gemma-3}). Adv. and Std. denote EX on the SAGE-generated adversarial evaluation set and the standard BIRD dev set, respectively.}
\label{tab:fine-tuning}
\vspace{-10pt}
\end{table}

\section{Conclusion}

In this paper, we propose \mname, a self-evolving framework for automated vulnerability discovery in Text-to-SQL. Extensive experiments confirm that \mname significantly outperforms static expert baselines in identifying weaknesses. We further demonstrate that the discovered vulnerabilities are highly transferable across different LLMs. Finally, our mitigation experiments show that \mname can effectively close the loop: the generated adversarial samples serve as valuable training data to repair model defects and improve overall robustness via lightweight fine-tuning.

\section*{Limitations}
We acknowledge two main limitations in our current work. First, our ablation studies are conducted primarily at the framework level; a more fine-grained analysis is required to isolate the specific contributions of internal components like the Judge, Embedding, and Summarizer modules. Second, the update mechanism of our Vulnerability Codex operates at coarse intervals; adopting a high-frequency, per-batch update strategy could further enhance the efficiency of the self-evolving process.

\bibliography{custom}

@inproceedings{zhu-etal-2025-fanno,
    title = "{FANNO}: Augmenting High-Quality Instruction Data with Open-Sourced {LLM}s Only",
    author = "Zhu, He  and
      Ding, Yifan  and
      Tao, Yicheng  and
      Ruan, Zhiwen  and
      Li, Yixia  and
      Zhang, Wenjia  and
      Chen, Yun  and
      Chen, Guanhua",
    editor = "Che, Wanxiang  and
      Nabende, Joyce  and
      Shutova, Ekaterina  and
      Pilehvar, Mohammad Taher",
    booktitle = "Findings of the Association for Computational Linguistics: ACL 2025",
    month = jul,
    year = "2025",
    address = "Vienna, Austria",
    publisher = "Association for Computational Linguistics",
    url = "https://aclanthology.org/2025.findings-acl.906/",
    doi = "10.18653/v1/2025.findings-acl.906",
    pages = "17633--17653",
    ISBN = "979-8-89176-256-5"
}

@inproceedings{zhu-etal-2025-tag,
    title = "Tag-Instruct: Controlled Instruction Complexity Enhancement through Structure-based Augmentation",
    author = "Zhu, He  and
      Ruan, Zhiwen  and
      Su, Junyou  and
      He, Xingwei  and
      Chen, Yun  and
      Zhang, Wenjia  and
      Chen, Guanhua",
    editor = "Che, Wanxiang  and
      Nabende, Joyce  and
      Shutova, Ekaterina  and
      Pilehvar, Mohammad Taher",
    booktitle = "Findings of the Association for Computational Linguistics: ACL 2025",
    month = jul,
    year = "2025",
    address = "Vienna, Austria",
    publisher = "Association for Computational Linguistics",
    url = "https://aclanthology.org/2025.findings-acl.911/",
    doi = "10.18653/v1/2025.findings-acl.911",
    pages = "17708--17729",
    ISBN = "979-8-89176-256-5"
}

@inproceedings{
lai2026biasscope,
title={BiasScope: Towards Automated Detection of Bias in {LLM}-as-a-Judge Evaluation},
author={Peng Lai and Zhihao Ou and Yong Wang and Longyue Wang and Jian Yang and Yun Chen and Guanhua Chen},
booktitle={The Fourteenth International Conference on Learning Representations},
year={2026},
url={https://openreview.net/forum?id=QGOw6AU8Lp}
}

@inproceedings{chi-etal-2025-pi,
    title = "Pi-{SQL}: Enhancing Text-to-{SQL} with Fine-Grained Guidance from Pivot Programming Languages",
    author = "Chi, Yongdong  and
      Wang, Hanqing  and
      Chen, Yun  and
      Yang, Yan  and
      Yang, Jian  and
      Yang, Zonghan  and
      Yan, Xiao  and
      Chen, Guanhua",
    editor = "Christodoulopoulos, Christos  and
      Chakraborty, Tanmoy  and
      Rose, Carolyn  and
      Peng, Violet",
    booktitle = "Findings of the Association for Computational Linguistics: EMNLP 2025",
    month = nov,
    year = "2025",
    address = "Suzhou, China",
    publisher = "Association for Computational Linguistics",
    url = "https://aclanthology.org/2025.findings-emnlp.1369/",
    doi = "10.18653/v1/2025.findings-emnlp.1369",
    pages = "25120--25144",
    ISBN = "979-8-89176-335-7"
}

@inproceedings{SpiderDK,
  title={Exploring Underexplored Limitations of Cross-Domain Text-to-SQL Generalization},
  author={Gan, Yujian and Chen, Xinyun and Purver, Matthew},
  booktitle={Proceedings of the 2021 Conference on Empirical Methods in Natural Language Processing},
  pages={8926--8931},
  year={2021}
}

@inproceedings{yu2018spider,
  title={Spider: A large-scale human-labeled dataset for complex and cross-domain semantic parsing and text-to-SQL task},
  author={Yu, Tao and Zhang, Rui and Yang, Kai and Yasunaga, Michihiro and Wang, Dongxu and Li, Zifan and Ma, James and Li, Irene and Yao, Qingning and Roman, Shanelle and others},
  booktitle={2018 Conference on Empirical Methods in Natural Language Processing, EMNLP 2018},
  pages={3911--3921},
  year={2018},
  organization={Association for Computational Linguistics}
}

@inproceedings{cheng2025sqlord,
  title={SQLord: A Robust Enterprise Text-to-SQL Solution via Reverse Data Generation and Workflow Decomposition},
  author={Cheng, Song and Cheng, Qiannan and Jin, Linbo and Yi, Lei and Zhang, Guannan},
  booktitle={Companion Proceedings of the ACM on Web Conference 2025},
  pages={919--923},
  year={2025}
}

@article{zhang2024benchmarking,
  title={Benchmarking the text-to-sql capability of large language models: A comprehensive evaluation},
  author={Zhang, Bin and Ye, Yuxiao and Du, Guoqing and Hu, Xiaoru and Li, Zhishuai and Yang, Sun and Liu, Chi Harold and Zhao, Rui and Li, Ziyue and Mao, Hangyu},
  journal={arXiv preprint arXiv:2403.02951},
  year={2024}
}

@article{team2025gemma,
  title={Gemma 3 technical report},
  author={Team, Gemma and Kamath, Aishwarya and Ferret, Johan and Pathak, Shreya and Vieillard, Nino and Merhej, Ramona and Perrin, Sarah and Matejovicova, Tatiana and Ram{\'e}, Alexandre and Rivi{\`e}re, Morgane and others},
  journal={arXiv preprint arXiv:2503.19786},
  year={2025}
}

@inproceedings{SpiderGen,
  title={Exploring dimensions of generalizability and few-shot transfer for text-to-SQL semantic parsing},
  author={Patil, Rajaswa and Patwardhan, Manasi and Karande, Shirish and Vig, Lovekesh and Shroff, Gautam},
  booktitle={Transfer Learning for Natural Language Processing Workshop},
  pages={103--114},
  year={2023},
  organization={PMLR}
}

@inproceedings{AmbiQT,
  title={Benchmarking and Improving Text-to-SQL Generation under Ambiguity},
  author={Bhaskar, Adithya and Tomar, Tushar and Sathe, Ashutosh and Sarawagi, Sunita},
  booktitle={Proceedings of the 2023 Conference on Empirical Methods in Natural Language Processing},
  pages={7053--7074},
  year={2023}
}

@article{saparina2024ambrosia,
  title={Ambrosia: A benchmark for parsing ambiguous questions into database queries},
  author={Saparina, Irina and Lapata, Mirella},
  journal={Advances in Neural Information Processing Systems},
  volume={37},
  pages={90600--90628},
  year={2024}
}

@inproceedings{SpiderSyn,
  title={Towards Robustness of Text-to-SQL Models against Synonym Substitution},
  author={Gan, Yujian and Chen, Xinyun and Huang, Qiuping and Purver, Matthew and Woodward, John R and Xie, Jinxia and Huang, Pengsheng},
  booktitle={Proceedings of the 59th Annual Meeting of the Association for Computational Linguistics and the 11th International Joint Conference on Natural Language Processing (Volume 1: Long Papers)},
  pages={2505--2515},
  year={2021}
}

@article{chang2023dr,
  title={Dr. spider: A diagnostic evaluation benchmark towards text-to-sql robustness},
  author={Chang, Shuaichen and Wang, Jun and Dong, Mingwen and Pan, Lin and Zhu, Henghui and Li, Alexander Hanbo and Lan, Wuwei and Zhang, Sheng and Jiang, Jiarong and Lilien, Joseph and others},
  journal={arXiv preprint arXiv:2301.08881},
  year={2023}
}

@inproceedings{ADVETA,
  title={Towards Robustness of Text-to-SQL Models Against Natural and Realistic Adversarial Table Perturbation},
  author={Pi, Xinyu and Wang, Bing and Gao, Yan and Guo, Jiaqi and Li, Zhoujun and Lou, Jian-Guang},
  booktitle={Proceedings of the 60th Annual Meeting of the Association for Computational Linguistics (Volume 1: Long Papers)},
  pages={2007--2022},
  year={2022}
}

@article{shen2025study,
  title={A Study of In-Context-Learning-Based Text-to-SQL Errors},
  author={Shen, Jiawei and Wan, Chengcheng and Qiao, Ruoyi and Zou, Jiazhen and Xu, Hang and Shao, Yuchen and Zhang, Yueling and Miao, Weikai and Pu, Geguang},
  journal={arXiv preprint arXiv:2501.09310},
  year={2025}
}

@article{ding2025ambisql,
  title={AmbiSQL: Interactive Ambiguity Detection and Resolution for Text-to-SQL},
  author={Ding, Zhongjun and Lin, Yin and Zeng, Tianjing},
  journal={arXiv preprint arXiv:2508.15276},
  year={2025}
}

@inproceedings{
gong2025sqlens,
title={{SQL}ens: An End-to-End Framework for Error Detection and Correction in Text-to-{SQL}},
author={Yue Gong and Chuan Lei and Xiao Qin and Kapil Vaidya and Balakrishnan Murali Narayanaswamy and Tim Kraska},
booktitle={The Thirty-ninth Annual Conference on Neural Information Processing Systems},
year={2025},
url={https://openreview.net/forum?id=on6Hf0KP20}
}

@article{hurst2024gpt,
  title={Gpt-4o system card},
  author={Hurst, Aaron and Lerer, Adam and Goucher, Adam P and Perelman, Adam and Ramesh, Aditya and Clark, Aidan and Ostrow, AJ and Welihinda, Akila and Hayes, Alan and Radford, Alec and others},
  journal={arXiv preprint arXiv:2410.21276},
  year={2024}
}

@article{androutsopoulos1995natural,
  title={Natural language interfaces to databases--an introduction},
  author={Androutsopoulos, Ion and Ritchie, Graeme D and Thanisch, Peter},
  journal={Natural language engineering},
  volume={1},
  number={1},
  pages={29--81},
  year={1995},
  publisher={Cambridge University Press}
}

@article{li2014constructing,
  title={Constructing an interactive natural language interface for relational databases},
  author={Li, Fei and Jagadish, Hosagrahar V},
  journal={Proceedings of the VLDB Endowment},
  volume={8},
  number={1},
  pages={73--84},
  year={2014},
  publisher={VLDB Endowment}
}

@article{li2024can,
  title={Can llm already serve as a database interface? a big bench for large-scale database grounded text-to-sqls},
  author={Li, Jinyang and Hui, Binyuan and Qu, Ge and Yang, Jiaxi and Li, Binhua and Li, Bowen and Wang, Bailin and Qin, Bowen and Geng, Ruiying and Huo, Nan and others},
  journal={Advances in Neural Information Processing Systems},
  volume={36},
  year={2024}
}

@article{liu2024epi,
  title={Epi-sql: Enhancing text-to-sql translation with error-prevention instructions},
  author={Liu, Xiping and Tan, Zhao},
  journal={arXiv preprint arXiv:2404.14453},
  year={2024}
}

@inproceedings{askari2025magic,
  title={Magic: Generating self-correction guideline for in-context text-to-sql},
  author={Askari, Arian and Poelitz, Christian and Tang, Xinye},
  booktitle={Proceedings of the AAAI Conference on Artificial Intelligence},
  volume={39},
  pages={23433--23441},
  year={2025}
}

@inproceedings{mahmud2015rule,
  title={A rule based approach for NLP based query processing},
  author={Mahmud, Tanzim and Hasan, KM Azharul and Ahmed, Mahtab and Chak, Thwoi Hla Ching},
  booktitle={2015 2nd international conference on electrical information and communication technologies (EICT)},
  pages={78--82},
  year={2015},
  organization={IEEE}
}

@article{lyu2020hybrid,
  title={Hybrid ranking network for text-to-sql},
  author={Lyu, Qin and Chakrabarti, Kaushik and Hathi, Shobhit and Kundu, Souvik and Zhang, Jianwen and Chen, Zheng},
  journal={arXiv preprint arXiv:2008.04759},
  year={2020}
}

@misc{infly2025infrlqwencoder,
  title        = {inf-rl-qwen-coder-32b-2746},
  author       = {Infly and InfTech},
  year         = {2025},
  publisher    = {Hugging Face},
  howpublished = {\url{https://huggingface.co/infly/inf-rl-qwen-coder-32b-2746}},
  note         = {Accessed: 2026-01-05}
}

@article{yang2025qwen3,
  title={Qwen3 technical report},
  author={Yang, An and Li, Anfeng and Yang, Baosong and Zhang, Beichen and Hui, Binyuan and Zheng, Bo and Yu, Bowen and Gao, Chang and Huang, Chengen and Lv, Chenxu and others},
  journal={arXiv preprint arXiv:2505.09388},
  year={2025}
}

@article{zhang2025qwen3,
  title={Qwen3 Embedding: Advancing Text Embedding and Reranking Through Foundation Models},
  author={Zhang, Yanzhao and Li, Mingxin and Long, Dingkun and Zhang, Xin and Lin, Huan and Yang, Baosong and Xie, Pengjun and Yang, An and Liu, Dayiheng and Lin, Junyang and others},
  journal={arXiv preprint arXiv:2506.05176},
  year={2025}
}

@inproceedings{
chang2023drspider,
title={Dr.Spider: A Diagnostic Evaluation Benchmark towards Text-to-{SQL} Robustness},
author={Shuaichen Chang and Jun Wang and Mingwen Dong and Lin Pan and Henghui Zhu and Alexander Hanbo Li and Wuwei Lan and Sheng Zhang and Jiarong Jiang and Joseph Lilien and Steve Ash and William Yang Wang and Zhiguo Wang and Vittorio Castelli and Patrick Ng and Bing Xiang},
booktitle={The Eleventh International Conference on Learning Representations },
year={2023},
url={https://openreview.net/forum?id=Wc5bmZZU9cy}
}

@article{choi2021ryansql,
  title={Ryansql: Recursively applying sketch-based slot fillings for complex text-to-sql in cross-domain databases},
  author={Choi, DongHyun and Shin, Myeong Cheol and Kim, EungGyun and Shin, Dong Ryeol},
  journal={Computational Linguistics},
  volume={47},
  number={2},
  pages={309--332},
  year={2021},
  publisher={MIT Press One Rogers Street, Cambridge, MA 02142-1209, USA journals-info~…}
}

@article{xu2017sqlnet,
  title={Sqlnet: Generating structured queries from natural language without reinforcement learning},
  author={Xu, Xiaojun and Liu, Chang and Song, Dawn},
  journal={arXiv preprint arXiv:1711.04436},
  year={2017}
}

@inproceedings{li2023resdsql,
  title={Resdsql: Decoupling schema linking and skeleton parsing for text-to-sql},
  author={Li, Haoyang and Zhang, Jing and Li, Cuiping and Chen, Hong},
  booktitle={Proceedings of the AAAI Conference on Artificial Intelligence},
  volume={37},
  pages={13067--13075},
  year={2023}
}

@inproceedings{yin2020tabert,
  title={TaBERT: Pretraining for Joint Understanding of Textual and Tabular Data},
  author={Yin, Pengcheng and Neubig, Graham and Yih, Wen-tau and Riedel, Sebastian},
  booktitle={Proceedings of the 58th Annual Meeting of the Association for Computational Linguistics},
  pages={8413--8426},
  year={2020}
}

@article{Survey,
author = {Shi, Liang and Tang, Zhengju and Zhang, Nan and Zhang, Xiaotong and Yang, Zhi},
title = {A Survey on Employing Large Language Models for Text-to-SQL Tasks},
year = {2025},
publisher = {Association for Computing Machinery},
address = {New York, NY, USA},
issn = {0360-0300},
url = {https://doi.org/10.1145/3737873},
doi = {10.1145/3737873},
note = {Just Accepted},
journal = {ACM Comput. Surv.},
month = jun
}

@article{pourreza2023din,
  title={Din-sql: Decomposed in-context learning of text-to-sql with self-correction},
  author={Pourreza, Mohammadreza and Rafiei, Davood},
  journal={Advances in Neural Information Processing Systems},
  volume={36},
  pages={36339--36348},
  year={2023}
}

@misc{xie2025opensearchsql,
      title={OpenSearch-SQL: Enhancing Text-to-SQL with Dynamic Few-shot and Consistency Alignment}, 
      author={Xiangjin Xie and Guangwei Xu and Lingyan Zhao and Ruijie Guo},
      year={2025},
      eprint={2502.14913},
      archivePrefix={arXiv},
      primaryClass={cs.CL},
      url={https://arxiv.org/abs/2502.14913}, 
}

@article{liu2025uncovering,
  title={Uncovering the Impact of Chain-of-Thought Reasoning for Direct Preference Optimization: Lessons from Text-to-SQL},
  author={Liu, Hanbing and Li, Haoyang and Zhang, Xiaokang and Chen, Ruotong and Xu, Haiyong and Tian, Tian and Qi, Qi and Zhang, Jing},
  journal={arXiv preprint arXiv:2502.11656},
  year={2025}
}

@inproceedings{xu2024chain,
  title={Chain-of-Program Prompting with Open-Source Large Language Models for Text-to-SQL},
  author={Xu, Bo and Li, Shufei and Wu, Yifei and Wei, Shouang and Du, Ming and Wang, Hongya and Song, Hui},
  booktitle={2024 International Joint Conference on Neural Networks (IJCNN)},
  pages={1--8},
  year={2024},
  organization={IEEE}
}

@article{xia2024r,
  title={$ r\hat{\ }3$:" This is My SQL, Are You With Me?" A Consensus-Based Multi-Agent System for Text-to-SQL Tasks},
  author={Xia, Hanchen and Jiang, Feng and Deng, Naihao and Wang, Cunxiang and Zhao, Guojiang and Mihalcea, Rada and Zhang, Yue},
  journal={arXiv preprint arXiv:2402.14851},
  year={2024}
}

@inproceedings{deng2025reforce,
  title={Reforce: A Text-to-SQL agent with self-refinement, format restriction, and column exploration},
  author={Deng, Minghang and Ramachandran, Ashwin and Xu, Canwen and Hu, Lanxiang and Yao, Zhewei and Datta, Anupam and Zhang, Hao},
  booktitle={ICLR 2025 Workshop: VerifAI: AI Verification in the Wild},
  year={2025}
}

@article{li2025omnisql,
  title={Omnisql: Synthesizing high-quality text-to-sql data at scale},
  author={Li, Haoyang and Wu, Shang and Zhang, Xiaokang and Huang, Xinmei and Zhang, Jing and Jiang, Fuxin and Wang, Shuai and Zhang, Tieying and Chen, Jianjun and Shi, Rui and others},
  journal={arXiv preprint arXiv:2503.02240},
  year={2025}
}

@inproceedings{qu-etal-2025-share,
    title = "{SHARE}: An {SLM}-based Hierarchical Action {C}or{RE}ction Assistant for Text-to-{SQL}",
    author = "Qu, Ge  and
      Li, Jinyang  and
      Qin, Bowen  and
      Li, Xiaolong  and
      Huo, Nan  and
      Ma, Chenhao  and
      Cheng, Reynold",
    editor = "Che, Wanxiang  and
      Nabende, Joyce  and
      Shutova, Ekaterina  and
      Pilehvar, Mohammad Taher",
    booktitle = "Proceedings of the 63rd Annual Meeting of the Association for Computational Linguistics (Volume 1: Long Papers)",
    month = jul,
    year = "2025",
    address = "Vienna, Austria",
    publisher = "Association for Computational Linguistics",
    url = "https://aclanthology.org/2025.acl-long.552/",
    doi = "10.18653/v1/2025.acl-long.552",
    pages = "11268--11292",
    ISBN = "979-8-89176-251-0"
}

@inproceedings{guo-etal-2025-sqlforge,
    title = "{SQLF}orge: Synthesizing Reliable and Diverse Data to Enhance Text-to-{SQL} Reasoning in {LLM}s",
    author = "Guo, Yu  and
      Jin, Dong  and
      Ye, Shenghao  and
      Chen, Shuangwu  and
      Jianyang, Jianyang  and
      Tan, Xiaobin",
    editor = "Che, Wanxiang  and
      Nabende, Joyce  and
      Shutova, Ekaterina  and
      Pilehvar, Mohammad Taher",
    booktitle = "Findings of the Association for Computational Linguistics: ACL 2025",
    month = jul,
    year = "2025",
    address = "Vienna, Austria",
    publisher = "Association for Computational Linguistics",
    url = "https://aclanthology.org/2025.findings-acl.443/",
    doi = "10.18653/v1/2025.findings-acl.443",
    pages = "8441--8452",
    ISBN = "979-8-89176-256-5"
}

@article{li2024codes,
  title={Codes: Towards building open-source language models for text-to-sql},
  author={Li, Haoyang and Zhang, Jing and Liu, Hanbing and Fan, Ju and Zhang, Xiaokang and Zhu, Jun and Wei, Renjie and Pan, Hongyan and Li, Cuiping and Chen, Hong},
  journal={Proceedings of the ACM on Management of Data},
  volume={2},
  number={3},
  pages={1--28},
  year={2024},
  publisher={ACM New York, NY, USA}
}

@article{sheng2025base,
  title={BASE-SQL: A powerful open source Text-To-SQL baseline approach},
  author={Sheng, Lei and Xu, Shuai-Shuai and Xie, Wei},
  journal={arXiv preprint arXiv:2502.10739},
  year={2025}
}

@inproceedings{deng2021structure,
  title={Structure-Grounded Pretraining for Text-to-SQL},
  author={Deng, Xiang and Hassan, Ahmed and Meek, Christopher and Polozov, Oleksandr and Sun, Huan and Richardson, Matthew},
  booktitle={Proceedings of the 2021 Conference of the North American Chapter of the Association for Computational Linguistics: Human Language Technologies},
  pages={1337--1350},
  year={2021}
}

@inproceedings{sahitaj2025utilising,
  title={Utilising large language models for adversarial attacks in text-to-sql: A perpetrator and victim approach},
  author={Sahitaj, Ariana and Nilles, Markus and Schenkel, Ralf and Schmitt, Vera},
  booktitle={Datenbanksysteme f{\"u}r Business, Technologie und Web (BTW 2025)},
  pages={919--931},
  year={2025},
  organization={Gesellschaft f{\"u}r Informatik, Bonn}
}

@article{li2023bird,
  title={BIRD: A big bench for large-scale database grounded text-to-SQL},
  author={Li, Jinyang and Hui, Binyuan and Qu, Ge and others},
  journal={NeurIPS},
  year={2023}
}

@article{hong2024next,
  title={Next-generation database interfaces: A survey of LLM-based text-to-SQL},
  author={Hong, Zijin and Yuan, Zheng and Zhang, Qinggang and others},
  year={2024},
  journal={arXiv preprint arXiv:2406.08426}
}

\appendix

\section{Complementary Experiments}
\subsection{Analysis of Discovered Vulnerability Patterns}\label{sec:error}

As illustrated in Figure~\ref{fig:error_category}, our framework uncovers a distinct ``Novel Patterns'' category that, while constituting only approximately $10\%$ to $11\%$ of the total error distribution, represents a critical set of previously overlooked vulnerabilities. Far from being a monolithic outlier, this category encapsulates a rich diversity of complex structural failures—most notably Schema Misinterpretation and Syntactic Overfitting—that distinctively differ from traditional error taxonomies. The significance of this segment extends far beyond its statistical proportion; as validated by our mitigation experiments, these specific failure modes serve as high-value training signals. The ability to leverage these patterns to construct effective data for model repair confirms that these newly identified vulnerabilities are not merely rare edge cases, but cornerstone defects whose resolution is fundamental to enhancing the overall robustness of LLMs.

\subsection{Cost-Effective Scalability on State-of-the-Art LLMs}
\label{sec:transfer-sota}

To assess the scalability of our approach while addressing the practical challenges of evaluating proprietary models, we targeted \textit{GPT-4o}. Motivated by the prohibitive inference costs associated with extensive adversarial optimization on such large-scale models, we adopted a budget-constrained evaluation protocol.

Instead of performing a full-scale search from scratch, we directly leveraged the vulnerability codex pre-distilled from the smaller \textit{Gemma-3} model. Furthermore, to strictly minimize computational overhead, we restricted the attack process on GPT-4o to a \textbf{single iteration}.

As presented in Table~\ref{tab:gpt4o_transfer}, despite this minimal investment of resources, our framework proved strikingly effective. The single-iteration attack successfully exposed significant vulnerabilities in GPT-4o, driving the Execution Accuracy down to $22.36\%$ and achieving a Vulnerability Exposure Rate (VER) of $65.80\%$. This result highlights a critical advantage of \mname: it enables researchers to uncover latent flaws in costly, state-of-the-art models by transferring ``knowledge'' from cheaper, open-source models, offering a highly resource-efficient pathway for robustness evaluation.

\begin{table}[t]
\centering
\small
\begin{tabular}{lccc}
\toprule
\textbf{Setting} & \textbf{EX} & \textbf{VER} & \textbf{ApD} \\
\midrule
Original & 65.38 & - & - \\
\addlinespace[0.2em]
\mname & \textbf{22.36} \textcolor{red}{(-43.02)} & 65.80 & 5.35 \\
\bottomrule
\end{tabular}
\caption{\textbf{Resource-efficient evaluation on GPT-4o.} Constrained by budget, we perform only a \textbf{single iteration} of attack using strategies transferred from \textit{Gemma-3}. Even under this restricted setting, the framework successfully identifies widespread vulnerabilities.}
\label{tab:gpt4o_transfer}
\end{table}

\section{More implementation details}
\subsection{Reproducibility Details}\label{sec:reproducibility}

Table~\ref{tab:reproducibility} consolidates the main search budget, inference settings, and the size of the initially-correct subset $|D|$ used to compute VER.

\begin{table*}[t]
\centering
\small
\begin{tabularx}{\textwidth}{lX}
\toprule
\textbf{Item} & \textbf{Details} \\
\midrule
Max iterations ($T$) & 3 \\
Hypotheses per sample ($K$) & 3 initial hypotheses per sample \\
Perturbation scopes & Query, Relevant Schema, and Irrelevant Schema \\
Experience retrieval & Top-5 entries retrieved from the current Vulnerability Codex \\
Similarity threshold ($\tau$) & 0.1 for semantic compression \\
Sampling settings & Default chat sampling settings of each backbone; for Qwen3-32B, we use $temperature=0.6$, $top\_k=20$, and $top\_p=0.95$ \\
Stopping rule & Stop searching a hypothesis once a valid failure is exposed or when the maximum iteration budget is reached \\
LLM modules & Generator / Checker / Summarizer: Qwen3-32B; Embedding model: Qwen3-Embedding-4B \\
Initially-correct subset $|D|$ on BIRD & Gemma-3: 823; Inf-rl-qwen: 1082; OmniSQL: 970 \\
Initially-correct subset $|D|$ on Spider & Gemma-3: 843; Inf-rl-qwen: 907; OmniSQL: 846 \\
\bottomrule
\end{tabularx}
\caption{Consolidated reproducibility details and the size of the initially-correct evaluation subset $|D|$ used to compute VER.}
\label{tab:reproducibility}
\end{table*}

\subsection{Detailed Algorithm of \mname}\label{sec:algorithm}
The complete Algorithm~\ref{alg:iterative_discovery_final} summarizes the implementation of \mname.
\begin{algorithm*}[t]
\centering
\caption{Systematic Automated Guided Exploration (SAGE)}
\label{alg:iterative_discovery_final}
\small
\begin{algorithmic}[1]
\Require Target Model $M_t$.
\Require Dataset $D = \{(q_i, s_i, y^*_i)\}_{i=1}^N$ where $M_t$ yields correct execution results for all $(q_i, s_i)$.
\Require LLM Modules: Generator $M_g$, Checker $M_c$, Summarizer $M_s$, Embedding Model $M_e$.
\Require Perturbation Scopes $\mathbb{S} = \{s_{\text{query}}, s_{\text{schema}}, \dots\}$.
\Ensure Refined Vulnerability Codex $\mathcal{V}_{final}$.

\State \textbf{Initialize:} Iteration counter $t \gets 1$, Vulnerability Codex $\mathcal{V}_0$.

\State \textbf{Phase 1: Systematic Hypothesis Generation}
\State Synthesize initial set of vulnerability hypotheses from dataset $D$:
\State $\mathcal{H}_{curr} \gets \bigcup_{d_i \in D} \{(d_i, h_{i,k}) \mid h_{i,k} \in \text{HypothesisGen}(M_g, d_i, K=3) \}$
\Comment{Each sample $d_i$ spawns $K$ distinct hypotheses}

\While{$t \leq T$ \textbf{and} $\mathcal{H}_{curr} \neq \emptyset$}
    \State $\mathcal{V}_{\text{new}}^{(t)} \gets \emptyset$ \Comment{Storage for new archetypes found in this iter}
    \State $\mathcal{H}_{next} \gets \emptyset$ \Comment{Hypotheses that fail to expose bugs are retained}
    
    \State \textbf{Phase 2: Experience-Guided Probing \& Verification}
    \For{each candidate tuple $C_{j}=(d_{i}, h_{i,k})$ in $\mathcal{H}_{curr}$}
        \State Retrieve relevant insights: $E_{j} \gets \text{RetrieveTopN}(M_e, \mathcal{V}_{t-1}, h_{i,k})$
        
        \State \Comment{Dynamic Scope Selection}
        \If{$t=1$}
            \State Select single scope: $\mathbb{S}_{active} \gets \{ \mathbb{S}[(k-1) \pmod{|\mathbb{S}|} + 1] \}$ 
            \Comment{Cold Start: Efficiently map 1 hypothesis to 1 scope}
        \Else
            \State Select all scopes: $\mathbb{S}_{active} \gets \mathbb{S}$ 
        \EndIf

        \State $IsExposed \gets \textbf{false}$

        \For{each scope $s_m \in \mathbb{S}_{active}$}
            \State \Comment{Apply hypothesis $h_{i,k}$ onto scope $s_m$ guided by experience $E_j$}
            \State $x_{j,m} \gets \text{SampleGenerate}(M_g, C_{j}, s_m, E_j)$
            
            \If{$\neg \text{QualityCheck}(x_{j,m}, M_c)$} \textbf{continue} \EndIf
            
            \State $y_{j,m} \gets \text{TargetModelQuery}(M_t, x_{j,m})$
            \State $IsCorrect \gets \text{VerifyAnswer}(d_{i}, x_{j,m}, y_{j,m})$
            
            \If{$\neg IsCorrect$} \Comment{Vulnerability Discovered}
                \State \Comment{Abstract the concrete failure case into a generalized archetype}
                \State $v_{j} \gets \text{AbstractVulnerability}(M_s, C_{j}, x_{j,m}, y_{j,m})$
                \If{$v_{j}$ is non-trivial}
                    \State $\mathcal{V}_{\text{new}}^{(t)} \gets \mathcal{V}_{\text{new}}^{(t)} \cup \{v_{\text{arch}}\}$
                \EndIf
                \State $IsExposed \gets \textbf{true}$
                \State \textbf{break} \Comment{Pruning: Hypothesis validated, skip remaining scopes}
            \EndIf
        \EndFor

        \If{$\neg IsExposed$}
            \State $\mathcal{H}_{next} \gets \mathcal{H}_{next} \cup \{C_j\}$ 
        \EndIf
    \EndFor

    \State \textbf{Phase 3: Codex Evolution \& Management}
    \State $\mathcal{V}_{t} \gets \mathcal{V}_{t-1} \cup \mathcal{V}_{\text{new}}^{(t)}$
    
    \State \Comment{Refine the Codex by merging similar archetypes}
    \State $\mathcal{V}_{t} \gets \text{SemanticCompress}(M_e, \mathcal{V}_{t}, \tau)$
    
    \State $\mathcal{H}_{curr} \gets \mathcal{H}_{next}$
    \State $t \gets t + 1$
    
\EndWhile

\State $\mathcal{V}_{final} \gets {V}_{t-1}$

\State \Return $\mathcal{V}_{final}$

\end{algorithmic}
\end{algorithm*}

\subsection{Sensitivity to Iteration Count}\label{sec:sensitivity_t}

To study the effect of the iteration budget, we evaluate \mname on Gemma-3 over the BIRD dev set while varying the maximum number of iterations $T$.

\begin{table}[t]
  \centering
  \small
  \begin{tabular}{lccc}
    \toprule
    \textbf{Iteration $T$} & \textbf{EX} & \textbf{VER} & \textbf{ApD} \\
    \midrule
    1 & 24.64 & 54.07 & 7.73 \\
    2 & 14.73 & 72.54 & 5.68 \\
    3 & 8.34 & 84.45 & 5.08 \\
    \bottomrule
  \end{tabular}
  \caption{Sensitivity of \mname to the iteration budget $T$ on Gemma-3 over the BIRD dev set.}
  \label{tab:sensitivity_t}
\end{table}

Increasing the iteration budget consistently improves VER while reducing ApD, indicating that a larger search budget uncovers more vulnerabilities and makes discovery more efficient.

\subsection{Training configurations of Fine-tuning Experiments}\label{sec:training_configs}

This section lists the key hyperparameters for the parameter-efficient fine-tuning experiments on \textit{Qwen2.5-Coder-7B-Instruct} in Table~\ref{tab:train_hyper}.

\begin{table}[t]
  \centering
  \small

    \begin{tabular}{m{3.5cm} m{3.5cm}}
    \toprule
    Hyperparameters & Details \\
    \midrule
    batch\_size & 64 \\
    Effective learning rate & $1.0\times10^{-4}$ \\
    Training epochs & 3 \\
    LR scheduler & cosine \\
    warmup & 0.1 \\
    Precision & bf16 \\
    Max seq length & 10240 \\
    \bottomrule
  \end{tabular}
  \caption{Training hyperparameters of fine-tuning experiments.}
  \label{tab:train_hyper}
\end{table}

\begin{figure*}[t]
    \centering
    \includegraphics[width=\textwidth]{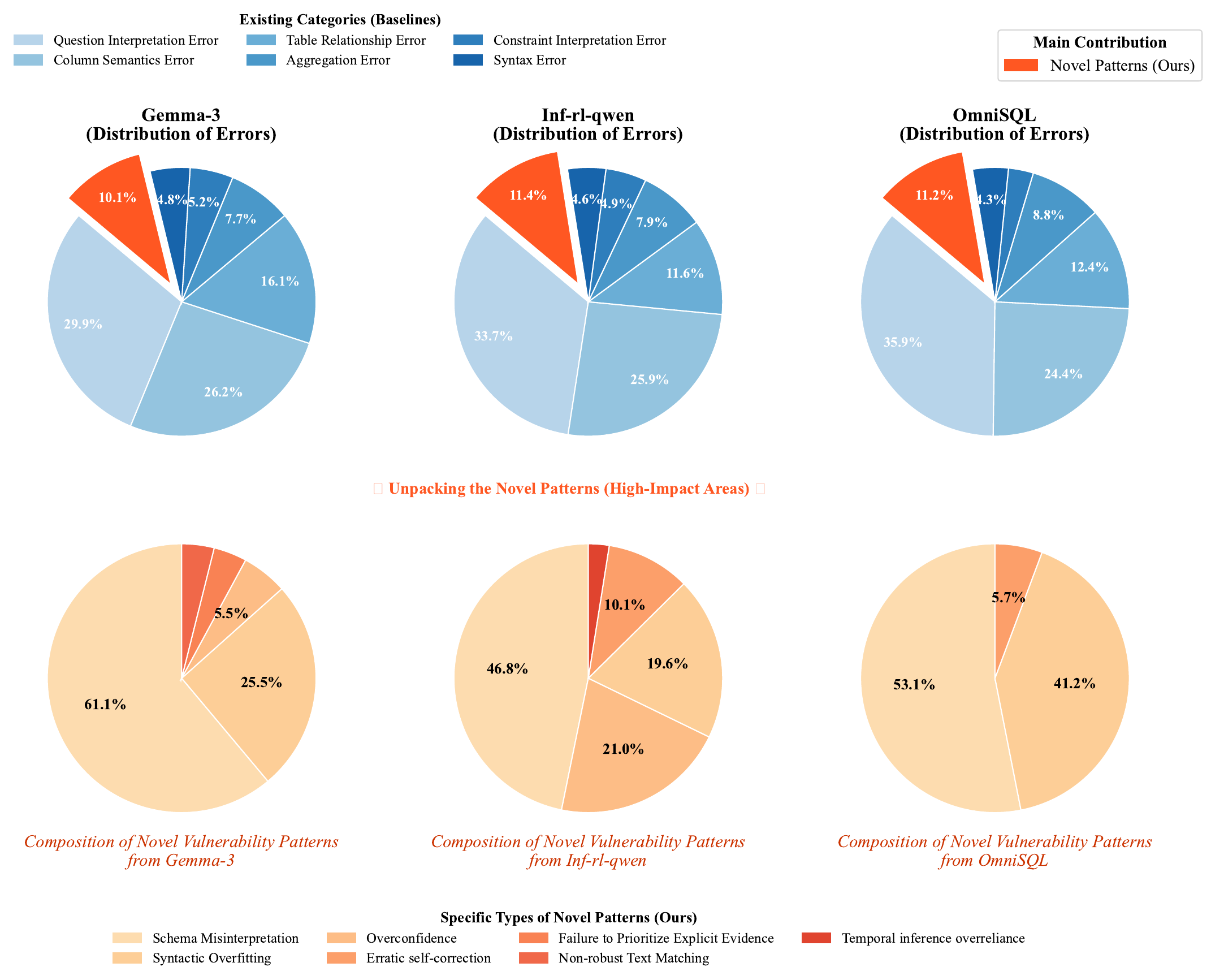}

    \caption{Distribution and hierarchical analysis of failure patterns across Gemma-3, Inf-rl-qwen, and OmniSQL. The top row illustrates the overall error distribution, distinguishing between established error categories (cool tones) and the Novel Patterns identified by our approach (orange, exploded slice). The bottom row provides a fine-grained breakdown of these novel patterns, revealing that specific failures—such as Schema Misinterpretation and Syntactic Overfitting—constitute a significant portion of errors previously categorized broadly as ``Other''.}

    \label{fig:error_category}
\end{figure*}

Beyond the established categories, Figure~\ref{fig:error_category} further presents a detailed analysis of errors typically grouped under the \textit{OTHER} category. By combining expert manual inspection with systematic reclassification, we uncover recurring and nuanced cognitive failure patterns. These include: schema misinterpretation, syntactic overfitting, insufficient prioritization of explicit evidence, overconfidence, non-robust text matching, erratic self-correction, and overreliance on temporal inference. This trigger-based partitioning moves beyond surface-level heuristics, substantially enriching standard error categories and, more importantly, providing fine-grained, actionable insights for targeted model improvement. 

\subsection{Prompt Templates for Sample Perturbation}
\label{sec:prompts}

We provide the prompt templates used to generate sample perturbations in our framework. Figures~\ref{fig:Q_prompts}, \ref{fig:IC_prompts}, and \ref{fig:RC_prompts} illustrate the prompt templates for perturbations applied to natural language queries, schema-irrelevant elements, and schema-relevant elements, respectively.

\begin{figure}[ht]
\small
\begin{tcolorbox}[enhanced, colback=white, title={The prompt templates for perturbations applied to natural language queries}]

\#\#\# Task:\\
Please **rewrite the question and the evidence** in a way that **preserves the original intent and does not change the final correct answer** (i.e., the gold SQL query must still be valid and applicable).\\
Your rewritten version should:\\
- Better reflect **real-world use cases or natural user phrasing**.\\
- Increase **linguistic and structural diversity** ...\\
- Introduce **more contextual framing, ambiguity, or reasoning cues** ...\\
- Remain **faithful to the original semantics and answer**.\\

\#\#\# Guidelines:\\
- Do **not** change the final answer (i.e., gold SQL must still be valid).\\
- Focus on **rewriting**, not expanding or omitting information.\\
- You must **describe your modification strategy** clearly in the `improvement` field.\\

\#\#\# Output Format:\\
Please return your output strictly in the following JSON format:\\
```json\\
\{\{\\
``improvement": ``Describe what changes you made and why ...",\\
  ``question": ``your rewritten question",\\
  ``evidence": ``your rewritten evidence"\\
\}\}\\
Example 1-3: ...\\

Now it is your turn:\\
Below is the database schema:\\
\{schema\}\\
Question: \{question\}\\
Evidence: \{evidence\}\\
The gold SQL query is: \{gold\_sql\}\\

\end{tcolorbox}
\caption{The prompt templates for perturbations applied to natural language queries.}
\label{fig:Q_prompts}
\end{figure}

\begin{figure}[ht]
\small
\begin{tcolorbox}[enhanced, colback=white, title={The prompt templates for perturbations applied to schema-irrelevant elements}]

\#\#\# Task:\\
Your task is to **modify the provided database-related information** while ensuring that the **core semantics and the correct answer remain unchanged**. \\
Your goal is to introduce **maximum lexical, structural, and semantic diversity** in order to **evaluate the model's robustness** against variations in metadata.\\
You are allowed to revise the following fields for each column:\\
- **column\_description**: Modify the column description using paraphrasing, synonyms, different sentence structures ...\\
- **value\_description**: Change the way value types, ranges, or categories are described ...\\
- **example\_value**: Add more diverse, realistic, or representative sample values ...\\

\#\#\# Important Guidelines:\\
- **Do not** change the table names or column names.\\
- **Do not** alter the underlying meaning or change the final answer.\\
- You must output your result in the **strict JSON format** shown below.\\
- You must change the metadata for **at least 3 columns** and **at most 5 columns**.\\

\#\#\# Output Format:\\
``table\_name": [\\
  \{\{\\
    ``column": ``column\_name",\\
    ``improvement": ``Describe your strategy for modifying this column's metadata.",\\
    ``column\_description": ``Modified description of the column.",\\
    ``value\_description": ``Modified description of the values.",\\
    ``example\_value": [``example1", ``example2"]\\
  \}\}\\

Example 1-3: ...\\

Below is a partial view of the database schema. It includes only the tables and columns that are directly relevant to answering the given question; all unrelated schema elements have been omitted.\\
\{json\_str\}\\
Question: \{question\}\\
Evidence: \{evidence\}\\
The gold SQL query is: \{gold\_sql\}\\
Here are some successful attack strategies you can refer to: \{error\_strategies\}\\
\# The column\_description and value\_description fields should each be no longer than 100 tokens.\\
Now it is your turn. Please return your answer strictly in the above JSON format.\\

\end{tcolorbox}
\caption{The prompt templates for perturbations applied to schema-irrelevant elements.}
\label{fig:IC_prompts}
\end{figure}

\begin{figure*}[ht]
\small
\begin{tcolorbox}[enhanced, colback=white, title={The prompt templates for perturbations applied to schema-relevant elements}]

You are an expert in prompt-based robustness evaluation for database-related tasks.

\#\#\# Task:\\
Please **modify the database-related information** while ensuring that the **core semantics and the correct answer remain unchanged**. Your goal is to introduce **greater lexical, structural, and semantic diversity** to challenge the robustness of the model. \\
You may revise the following elements for each column:\\
- **column\_description**: Modify the description using paraphrasing, synonyms, different sentence structures, or multilingual expressions.\\
- **value\_description**: Alter the value type description or re-express the value mappings, ranges, or classifications.\\
- **example\_value**: Enrich the example values with varied, realistic, and semantically appropriate entries.\\

**Important:**\\
- Do **not** change the table name or column name.\\
- Your output must be in **strict JSON format** as shown below.\\
- You must change the metadata for **at least 3 columns** and **at most 5 columns**.\\

Here are some output examples:\\
Example 1-3: ...\\
Now it is your turn.\\
Below is a partial view of the database schema. It includes only the tables and columns that are relevant to answering the given question; unrelated schema elements have been omitted.\\
\{json\_str\}\\
Question: \{question\}\\
Evidence: \{evidence\}\\
The gold SQL query is: \{gold\_sql\}\\

Return your modifications using the following structure:\\
\{\{\\
  ``table\_name": [\\
    \{\{\\
      ``column": ``column\_name",\\
      ``improvement": ``Describe your strategy for modifying this column's metadata.",\\
      ``column\_description": ``Modified description of the column.",\\
      ``value\_description": ``Modified description of the values.",\\
      ``example\_value": [``example1", ``example2"]\\
    \}\}\\
  ]\\
\}\}\\

Here are some successful attack strategies you can refer to: \{error\_strategies\}\\
The column\_description and value\_description fields should each be no longer than 100 tokens.\\
Please return your answer strictly in the above JSON format.\\

\end{tcolorbox}
\caption{The prompt templates for perturbations applied to schema-relevant elements.}
\label{fig:RC_prompts}
\end{figure*}

\subsection{Checker validation}\label{sec:checker_validation}

To verify the reliability of the automated Checker (implemented with Qwen3-32B), we conducted both a targeted human evaluation and a larger-scale cross-model proxy validation. We recruited three graduate-level annotators with expertise in Text-to-SQL tasks to ensure the quality of the assessment. They were compensated at a competitive market rate for such tasks. All participants were informed about the purpose of the evaluation task and how the data would be used prior to the annotation. They provided informed consent to participate in the study. Since the study involves standard data annotation tasks with minimal risk to participants and does not collect personally identifiable information (PII), it was determined to be exempt from formal IRB review according to our institution's guidelines.

From the Gemma-3 experiments, we randomly sampled 100 perturbed cases for evaluation. Each annotator independently judged whether each perturbed query was valid (i.e., the perturbation preserves semantic equivalence and the correct answer should remain unchanged) or invalid (i.e., the perturbation alters the meaning or introduces inconsistencies).

Each annotator was provided with (i) the original natural-language query and its gold answer, (ii) the perturbed query, and (iii) the corresponding M-schema context. They followed a concise guideline: (1) assess semantic equivalence between the original and perturbed queries, (2) verify whether the perturbation is realistic and contextually appropriate, and (3) label the sample as either Valid or Invalid. The final ground-truth label for each case was determined by majority vote across the three annotators to reduce subjectivity and ensure consistency.

We then compared the Checker’s predictions with these human consensus labels. The Checker achieved an average accuracy of 89.3\%, indicating a high level of agreement with expert human judgments in identifying valid perturbations.

To complement this targeted human assessment, we further sampled 2,000 perturbations that were originally validated by the Qwen3-32B Checker and re-evaluated them using GPT-5.1 as an independent proxy judge. Table~\ref{tab:checker_proxy_validation} reports the agreement rates across different perturbation scopes.

\begin{table}[t]
  \centering
  \small
  \begin{tabular}{lcc}
    \toprule
    \textbf{Scope} & \textbf{\# Samples} & \textbf{Agreement} \\
    \midrule
    Query Only & 661 & 96.52 \\
    Irrelevant Schema & 662 & 95.62 \\
    Relevant Schema & 677 & 87.44 \\
    Overall & 2000 & 93.15 \\
    \bottomrule
  \end{tabular}
  \caption{Cross-model proxy validation of the Qwen3-32B Checker using GPT-5.1 as an independent judge. Agreement is reported in percentage.}
  \label{tab:checker_proxy_validation}
\end{table}

The Relevant Schema setting is the most challenging because modifying metadata for columns used in the gold SQL more easily introduces subtle semantic ambiguity. We therefore view this proxy study as complementary evidence rather than a replacement for human judgment. At the same time, the consistently high agreement and the mitigation gains in Table~\ref{tab:fine-tuning} jointly suggest that the discovered perturbations are not dominated by invalid rewrites.

\section{Ethical Considerations}
This work aims to improve the robustness of Text-to-SQL systems by automating the discovery of corner cases and vulnerabilities. Here, ``vulnerability'' refers to robustness and reliability brittleness under semantic-preserving perturbations, not SQL-injection-style exploits. While our method is designed to facilitate debugging and model improvement, we recognize that it carries a potential risk of misuse for adversarial attacks. We emphasize that identifying these weaknesses proactively is essential for mitigating risks in real-world deployments, and we urge users to apply this framework responsibly to enhance system security.

\end{document}